\documentclass[lettersize, journal]{IEEEtran}

\usepackage[colorlinks,linkcolor=blue,citecolor=blue]{hyperref}
\usepackage{graphicx}
\usepackage{epstopdf}
\usepackage{tabularray}
\usepackage{booktabs}
\usepackage{bm}
\usepackage{indentfirst}
\usepackage{float}
\usepackage{CJK}
\usepackage{pifont}
\usepackage{amsmath,amsfonts}
\usepackage{algorithmic}
\usepackage{algorithm}

\usepackage{array}
\usepackage[caption=false,font=normalsize,labelfont=sf,textfont=sf]{subfig}
\usepackage{textcomp}
\usepackage{stfloats}
\usepackage{url}
\usepackage{verbatim}
\usepackage{balance}
\usepackage{multirow}
\usepackage[normalem]{ulem}
\useunder{\uline}{\ul}{}
\usepackage{threeparttable}
\usepackage{tikz,orcidlink}

\def\BibTeX{{\rm B\kern-.05em{\sc i\kern-.025em b}\kern-.08em
    T\kern-.1667em\lower.7ex\hbox{E}\kern-.125emX}}
\usepackage{balance}
\begin{document}

\title{FTDMamba: Frequency-Assisted Temporal Dilation Mamba for Unmanned Aerial Vehicle Video Anomaly Detection}

\author{Cheng-Zhuang Liu$^{~\orcidlink{0009-0007-5413-6850}}$,
Si-Bao Chen$^{~\orcidlink{0000-0003-1481-0162}}$, ~\IEEEmembership{Member, ~IEEE}, Qing-Ling Shu$^{~\orcidlink{0009-0003-1868-4498}}$, Chris Ding$^{~\orcidlink{0009-0009-3374-1941}}$, \\ 
Jin Tang$^{~\orcidlink{0000-0001-8375-3590}}$, Bin Luo$^{~\orcidlink{0000-0001-5948-5055}}$, ~\IEEEmembership{Senior~Member, ~IEEE} 
\thanks{
This work was supported in part by the NSFC Key Project of Joint Fund for Enterprise Innovation and Development under Grant U24A20342, and in part by the National Natural Science Foundation of China under Grant 62576006 and 61976004.  (Corresponding author: Si-Bao Chen.) }
\thanks{
Cheng-Zhuang Liu, Si-Bao Chen, Qing-Ling Shu, Jin Tang and Bin Luo are affiliated with Information Materials and Intelligent Sensing Laboratory of Anhui Province, Anhui Provincial Key Laboratory of Multimodal Cognitive Computation, School of Computer Science and Technology, Anhui University, Hefei 230601, China (e-mail: chengzhuangliu@163.com; sbchen@ahu.edu.cn; 2563489133@qq.com; tj@ahu.edu.cn; luobin@ahu.edu.cn). 

Chris Ding is with the Department of Computer Science and
Engineering, The Chinese University of Hong Kong, Shenzhen 518172, China
(e-mail: chrisding@cuhk.edu.cn).
}}

\maketitle
\begin{abstract}
Recent advances in video anomaly detection (VAD) mainly focus on ground-based surveillance or unmanned aerial vehicle (UAV) videos with static backgrounds, whereas research on UAV videos with dynamic backgrounds remains limited.
Unlike static scenarios, dynamically captured UAV videos exhibit multi-source motion coupling, where the motion of objects and UAV-induced global motion are intricately intertwined.
Consequently, existing methods may misclassify normal UAV movements as anomalies or fail to capture true anomalies concealed within dynamic backgrounds. Moreover, many approaches do not adequately address the joint modeling of inter-frame continuity and local spatial correlations across diverse temporal scales.
To overcome these limitations, we propose the Frequency-Assisted Temporal Dilation Mamba (FTDMamba) network for UAV VAD, including two core components: (1) a Frequency Decoupled Spatiotemporal Correlation Module, which disentangles coupled motion patterns and models global spatiotemporal dependencies through frequency analysis; and (2) a Temporal Dilation Mamba Module, which leverages Mamba's sequence modeling capability to jointly learn fine-grained temporal dynamics and local spatial structures across multiple temporal receptive fields. 
Additionally, unlike existing UAV VAD datasets which focus on static backgrounds, we construct a large-scale Moving UAV VAD dataset (MUVAD), comprising 222,736 frames with 240 anomaly events across 12 anomaly types.
Extensive experiments demonstrate that FTDMamba achieves state-of-the-art (SOTA) performance on two public static benchmarks and the new MUVAD dataset. The code and MUVAD dataset will be available at: \url{https://github.com/uavano/FTDMamba}.
\end{abstract}


\vspace{5pt}

\section{Introduction}
Video anomaly detection (VAD) constitutes a fundamental technology in intelligent surveillance systems, aimed at automatically identifying events or behaviors that deviate from normal patterns within video sequences~\cite{huang2021abnormal}. Due to the rarity and diversity of anomalous events~\cite{10097199}, obtaining sufficient anomaly samples is particularly challenging. Consequently, unsupervised one-class classification (OCC) VAD methods~\cite{ramachandra2020survey,liu2018future,liu2024vadiffusion,10794594} have become the dominant paradigm. These methods train a model using only normal data to learn the distribution of normal events, subsequently identifying data that significantly deviates from this distribution as anomalies.

\begin{figure}[t]
	\centering
	\includegraphics[width=0.45\textwidth,page=1]{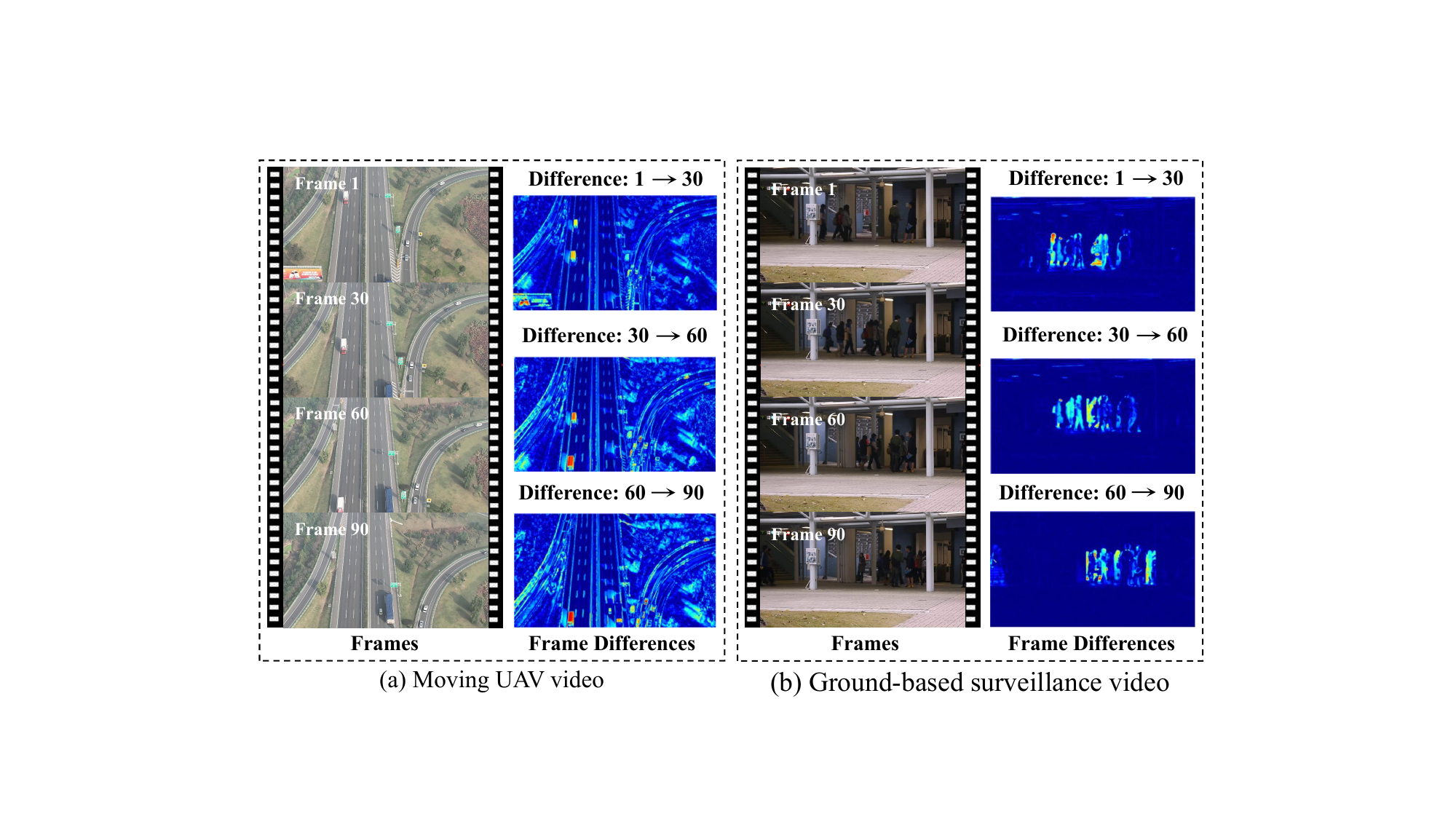} 
	\caption{
Illustration of the multi-source motion coupling issue. (a) and (b) show the frame differences between UAV videos with dynamic backgrounds and ground-based surveillance videos with static backgrounds.
	}      
	\label{Figure1}
\end{figure}

With the rapid advancement of unmanned aerial vehicle (UAV) technology and its widespread deployment across urban security~\cite{10753337}, traffic management~\cite{zhou2020resilient}, and environmental monitoring~\cite{guan2019novel} domains, UAV-based VAD has demonstrated substantial application potential~\cite{jin2022anomaly,tran2024transformer,le2025hstforu}. It is noteworthy that practical UAV deployments typically require continuous motion rather than stationary hovering. This  introduces a fundamental technical challenge of multi-source motion coupling. Specifically, the global background flow induced by the UAV's ego-motion is entangled with the local motion of foreground objects, causing the observed motion vector of any given pixel to be the vector superposition of these two independent components.
As illustrated in Fig.  \ref{Figure1}, the frame differences indicate that motion variations in static background videos primarily originate from foreground objects. In contrast, motion variations in dynamic UAV videos result from the combined effects of local object motion and global background motion. Therefore, existing VAD methods~\cite{ahn2024videopatchcore,cai2021appearance,chang2022video,9622181,wang2023memory} designed for static backgrounds encounter significant challenges when applied to dynamic UAV scenarios.

\begin{figure}[t]
	\centering
	\includegraphics[width=0.45\textwidth,page=1]{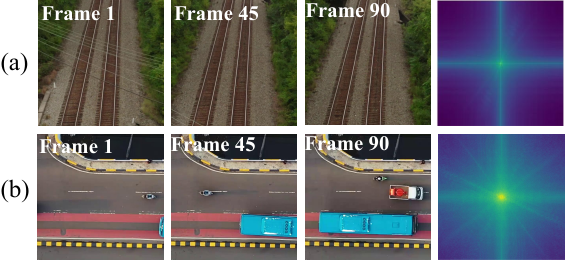} 
	\caption{
    Comparison of spectral distributions for global and local motion patterns. (a) Pure global motion: average frequency spectrum computed from 30 frames containing only UAV-induced motion. (b) Pure local motion: average frequency spectrum computed from 30 frames with static backgrounds and exclusively moving objects. 
	}      
	\label{Figure2}
\end{figure}

Firstly, existing VAD strategies are generally classified into reconstruction-based~\cite{hasan2016learning,gong2019memorizing,park2020learning,chang2020clustering} and prediction-based~\cite{le2023attention,qiu2024video,zhao2025rethinking,chen2022comprehensive}  approaches. Both paradigms mainly learn normal patterns through spatiotemporal modeling and identify anomalies as regions with large reconstruction or prediction errors. However, these methods extract only entangled motion features, lacking the ability to distinguish background from foreground motion. As a result, normal UAV movements may be misclassified as anomalies, while true abnormal events are obscured by dynamic backgrounds.
Given the inherent limitations of spatiotemporal VAD methods, frequency domain analysis emerges as a promising way. 
The core motivation is that global and local motion patterns exhibit distinct spectral characteristics~\cite{bex2002comparison,park2025spectral}. As illustrated in Fig.  \ref{Figure2}, the first row shows the average spectrum of pure global motion, with energy highly concentrated along cross-shaped axes through the spectrum center, reflecting the structured characteristics of background motion. The second row presents the average spectrum of pure local motion, where energy disperses more broadly without the distinctive cross-structure, indicating the unstructured nature of spatially localized object motion.
The most related VAD work about frequency is FE-VAD~\cite{pi2024fe}. FE-VAD sequentially applies Fourier transforms~\cite{duhamel1990fast} across temporal and spatial dimensions to decouple features into high and low frequency branches, enhancing anomaly detection through weighted spatio-temporal feature integration from both frequency bands. However, FE-VAD has two key limitations: (1) it employs fixed Gaussian filters for frequency separation, which cannot adapt to complex motion coupling in dynamic scenes; (2) it sequentially processes temporal and spatial dimensions, overlooking dependencies among frequency components across different spatiotemporal locations.

Secondly, the motion characteristics of UAV videos generally exhibit distinct multi-scale temporal patterns. Specifically, UAV global motion typically manifests as continuous and relatively smooth patterns with slow and stable changes, requiring a larger temporal window for modeling. In contrast, local object motion demonstrates diverse and irregular patterns, with more abrupt changes that necessitate shorter temporal windows to capture fine-grained dynamic information. Although some existing methods~\cite{zhang2024multi,zhong2022bidirectional,wang2021robust} include multi-scale modeling, they mainly focus on different object scales in the spatial dimension rather than multi-scale features in the temporal dimension. As a result, existing approaches insufficiently exploit spatiotemporal features across multiple temporal resolutions.

Based on the preceding analysis, this paper proposes a novel Frequency-Assisted Temporal Dilation Mamba (FTDMamba) network for dynamic UAV VAD, which integrates complementary frequency-domain and spatiotemporal perspectives.
To tackle the multi-source motion coupling challenge, we propose the Frequency Decoupled Spatiotemporal Correlation Module. This module first employs 1D Fast Fourier Transform (FFT)~\cite{yin2019fourier} along the temporal dimension to decompose coupled motion signals into distinct frequency bands, thereby achieving preliminary motion decoupling. Subsequently, it computes spatiotemporal autocorrelation matrices via 2D FFT in the spatiotemporal domain, which serve as attention weights to model global spatiotemporal dependencies.
To jointly capture multi-scale temporal continuity and local spatial correlations, we design the Temporal Dilation Mamba Module. Leveraging Mamba's~\cite{gu2023mamba} sequence modeling capabilities, this module constructs two complementary scanning sequences: pixel-level temporal-first scanning to capture fine-grained temporal dynamics, and patch-level spatial-first scanning to model local spatial structures. Furthermore, it incorporates a temporal dilation modeling strategy that provides diverse temporal receptive fields to capture multi-level motion patterns.

In addition, existing public UAV video datasets, such as Drone-Anomaly~\cite{jin2022anomaly} and UIT-ADrone~\cite{tran2023uit}, are mainly collected from static hovering drones. However, in practical scenarios, UAVs are generally in motion, causing anomalous behaviors that can be accurately detected under static backgrounds exhibit significantly degraded detection performance in dynamic backgrounds. To bridge the gap between existing datasets and practical deployment requirements, we construct a large-scale moving UAV VAD dataset (MUVAD). MUVAD includes 46 training and 72 testing video sequences, totaling 222,736 frames with 240 anomaly events  across 12 anomaly types. MUVAD provides realistic moving UAV video data, advancing the real-world application of UAV VAD.

Our main contributions can be summarized as follows:
\begin{itemize}
    \item We construct a large-scale dataset MUVAD captured by moving UAVs to better simulate real-world application scenarios, filling the gap in existing datasets and providing crucial support for in-depth research in UAV VAD.
    \item We propose a novel Frequency-Assisted Temporal Dilation Mamba framework for dynamic UAV VAD, comprising two core modules: Frequency Decoupled Spatiotemporal Correlation module and Temporal Dilation Mamba module. The former decouples  coupled motion in UAV videos and models global spatiotemporal relations. The latter jointly models the temporal continuity and local spatial correlations across multiple temporal scales. 
    \item We conduct extensive experiments on two public benchmarks and our newly constructed MUVAD dataset. Our method achieves SOTA performance across all three datasets, demonstrating its superior capability in handling both static and dynamic background scenarios. This validates the effectiveness and generalization of our method.
\end{itemize}

\section{Related Work}

\subsection{Unsupervised Spatiotemporal VAD}
VAD is a key research problem in computer vision, which has long been focused on videos with static backgrounds. Most existing unsupervised VAD methods perform modeling in the spatiotemporal domain, and they can be broadly categorized into ground-based and UAV-based VAD methods.

\subsubsection{Ground-based VAD methods}
Ground-based VAD methods have been extensively studied and can be mainly categorized into reconstruction-based and prediction-based methods.

The reconstruction-based methods assume that models trained to reconstruct normal video frames perform poorly in reconstructing abnormal frames. Hasan \emph{et al}.~\cite{hasan2016learning} are the first to propose using convolutional autoencoders for VAD, learning the representation of normal patterns by minimizing the reconstruction error of normal video frames. Gong \emph{et al}.~\cite{gong2019memorizing} propose the memory-augmented autoencoder, which introduces a memory module to store prototype representations of normal patterns, effectively improving the accuracy of anomaly detection. MNAD~\cite{park2020learning} further improves the memory mechanism by proposing an attention-based memory network that can adaptively select relevant memory items for reconstruction. Wang \emph{et al}.~\cite{chang2020clustering} integrate deep k-means clustering with autoencoders to map normal samples near cluster centers and separate anomalies by capturing variation factors. 

The prediction-based methods assume models trained to predict future frames of normal video and cannot accurately predict abnormal frames. Liu \emph{et al}.~\cite{liu2018future} first propose a prediction-based VAD framework that identifies anomalies by comparing predicted future frames with actual frames. Wang \emph{et al}.~\cite{chen2022comprehensive} propose a bidirectional architecture for VAD, comprehensively regularizing the prediction task through three constraints: prediction consistency, correlation consistency, and temporal consistency. Besides, some studies~\cite{cao2024scene, liu2021hybrid,yu2022abnormal} enhance the accuracy of VAD by integrating optical flow or inter-frame difference information. Furthermore, Yang \emph{et al}.~\cite{yang2023video} incorporate keyframes with implicit appearance and motion cues into VAD to infer missing intermediate frames for video event restoration. MA-PDM~\cite{zhou2025video} proposes a patch-based diffusion model  by leveraging appearance and motion conditioning. DoTA~\cite{yao2022dota} introduces future object localization technology into VAD, identifying anomalies by evaluating the accuracy and consistency of predicted future object positions. 

Reconstruction-based methods are inherently limited in modeling long-term temporal dependencies, hindering further performance gains. Therefore, prediction-based VAD frameworks have become the prevailing paradigm. In this paper, we formulate our approach within this predictive paradigm.

\subsubsection{UAV-based VAD Methods}
Compared to ground-based surveillance videos, UAV videos have distinctive characteristics such as varying viewpoints, complex motion. These features bring new technical challenges to UAV VAD. 
ANDT~\cite{jin2022anomaly} first applies Transformer architecture to UAV VAD by representing video sequences as spatiotemporal tubelet embeddings and leveraging the Transformer's global receptive field for future frame prediction.
ASTT~\cite{tran2024transformer} employs separate spatial and temporal Transformer encoders to learn frame spatial features and inter-frame temporal dependencies,  subsequently fusing them with a cross-attention mechanism to predict future frames.
Following the prediction paradigm, HSTforU~\cite{le2025hstforu} integrates a joint spatio-temporal attention mechanism into the U-Net framework to model the spatio-temporal interactions within multi-scale feature maps.

Existing ground-based or UAV-based VAD methods generally conduct unified spatiotemporal modeling of complex motion patterns. However, in dynamic UAV videos, they fail to distinguish UAV-induced background variations from true anomalies. To address this, we complement spatiotemporal modeling with frequency domain motion decoupling.

\subsection{Mamba in Computer Vision}
The structured state space models S4~\cite{gu2022efficiently} efficiently model long sequences by discretizing continuous systems, but their fixed parameters limit content awareness. Mamba~\cite{gu2023mamba} overcomes this limitation with a selectivity mechanism to achieve input-dependent dynamic parameter tuning. Following its success in NLP, Mamba has been extended to computer vision. Vision Mamba~\cite{zhu2024vision} and VMamba~\cite{liu2024vmamba} apply Mamba to image processing using patch serialization and cross-scanning, while VideoMamba~\cite{li2024videomamba} extends this to video understanding by integrating 3D selective scanning and temporal modeling. In the anomaly detection field, MambaAD~\cite{he2024mambaad} combines Mamba’s global modeling with CNN’s local feature extraction for industrial anomaly detection.  STNMamba~\cite{li2024stnmamba} extracts features from both the original videos and frame differences using Mamba and unifies them into the same feature space.  VADMamba~\cite{lyu2025vadmamba} leverages the Mamba to extract features from original videos and optical flows, accurately detecting anomalies by integrating frame prediction errors and optical flow reconstruction errors.
Inspired by Mamba’s sequence modeling, we propose temporal-first and spatial-first scanning sequences across different temporal receptive fields to model temporal continuity and local spatial correlations.

\section{Preliminaries}

\subsection{Wiener-Khinchin Theorem}
The Wiener-Khinchin theorem~\cite{cohen1998generalization} establishes a fundamental relationship between the autocorrelation function of a signal and its power spectral density (PSD). For a signal $x[n]$, the autocorrelation function $R_{xx}[\tau]$ measures the similarity between the signal and a time-shifted version of itself:
\begin{equation}
	\begin{gathered}
		R_{xx}[\tau] = \mathbb{E}[x[n] \cdot x[n+\tau]],
	\end{gathered}
	\label{eq7}
\end{equation}
where $\mathbb{E}[\cdot]$ denotes the expected value and $\tau$ is the time lag. The theorem states that the autocorrelation function and the PSD form a Fourier transform pair:
\begin{equation}
	\begin{gathered}
		S_{xx}(f) = \mathcal{F}\{R_{xx}[\tau]\} = \sum_{\tau=-\infty}^{\infty} R_{xx}[\tau] e^{-j2\pi f\tau},
	\end{gathered}
	\label{eq8}
\end{equation}
\begin{equation}
	\begin{gathered}
		R_{xx}[\tau] = \mathcal{F}^{-1}\{S_{xx}(f)\} = \int_{-\infty}^{\infty} S_{xx}(f) e^{j2\pi f\tau} df,
	\end{gathered}
	\label{eq9}
\end{equation}
where $\mathcal{F}\{\cdot\}$ and $\mathcal{F}^{-1}\{\cdot\}$ denote the Fourier transform and inverse Fourier transform, respectively. In practice, the PSD can be efficiently computed via FFT:
\begin{equation}
	\begin{gathered}
		S_{xx}(f) = X(f) \cdot X^*(f) = |X(f)|^2,
	\end{gathered}
	\label{eq10}
\end{equation}
where $X(f) = \mathcal{FFT}\{x[n]\}$, $X^*(f)$ is its complex conjugate, and $|\cdot|$ is the magnitude. The autocorrelation function is then obtained by applying the inverse FFT: $R_{xx}[\tau] = \mathcal{FFT}^{-1}\{S_{xx}(f)\}$. This theorem is powerful for video analysis, enabling efficient computation of spatiotemporal correlations in the frequency domain, thus avoiding costly direct convolution in the spatiotemporal domain.

\subsection{State Space Models (SSMs)}
SSMs are commonly considered as linear time-invariant systems that map a 1-D sequence $x(t) \in \mathbb{R} \mapsto y(t) \in \mathbb{R}$ through a hidden state $h(t) \in \mathbb{R}^N$. This system is typically formulated as linear ordinary differential equations (ODEs):
\begin{equation}
	\begin{gathered}
		h'(t) = A h(t) + B x(t), \quad y(t) = C h(t),
	\end{gathered}
	\label{eq1}
\end{equation}
where $A \in \mathbb{R}^{N \times N}$ is the transition state matrix, $B \in \mathbb{R}^{N \times 1}$ and $C \in \mathbb{R}^{1 \times N}$ are the projection matrices.

The structured SSMs $(S_4)$ converts the continuous parameters $A$ and $B$ into discrete parameters $\bar{A}$ and $\bar{B}$ by introducing a temporal scaling parameter $\Delta$ (Zero-order Hold):
\begin{equation}
	\begin{gathered}
		\bar{A} = \exp(\Delta A), \quad \bar{B} = (\Delta A)^{-1} (\exp(\Delta A) - I) \Delta B.
	\end{gathered}
	\label{eq2}
\end{equation}
After $\bar{A}$ and $\bar{B}$ are discretized, the system equations can be reformulated with a step size $\Delta$ as follows:
\begin{equation}
	\begin{gathered}
		h_t = \bar{A} h_{t-1} + \bar{B} x_t, \quad y_t = C h_t.
	\end{gathered}
	\label{eq3}
\end{equation}
The models ultimately compute outputs through a global convolution operation:
\begin{equation}
	\begin{gathered}
		\bar{K} = ({C}\bar{B}, {C}\bar{A}\bar{B}, {C}\bar{A}^2\bar{B},  {C}\bar{A}^{M-1}\bar{B}), \quad y = x * \bar{K},
	\end{gathered}
	\label{eq4}
\end{equation}
where $M$ denotes the length of the input sequence $x$, and $\bar{K} \in \mathbb{R}^M$ represents a structured convolutional kernel.

\begin{figure*}[t]
	\centering
	\includegraphics[width=0.95\textwidth,page=1]{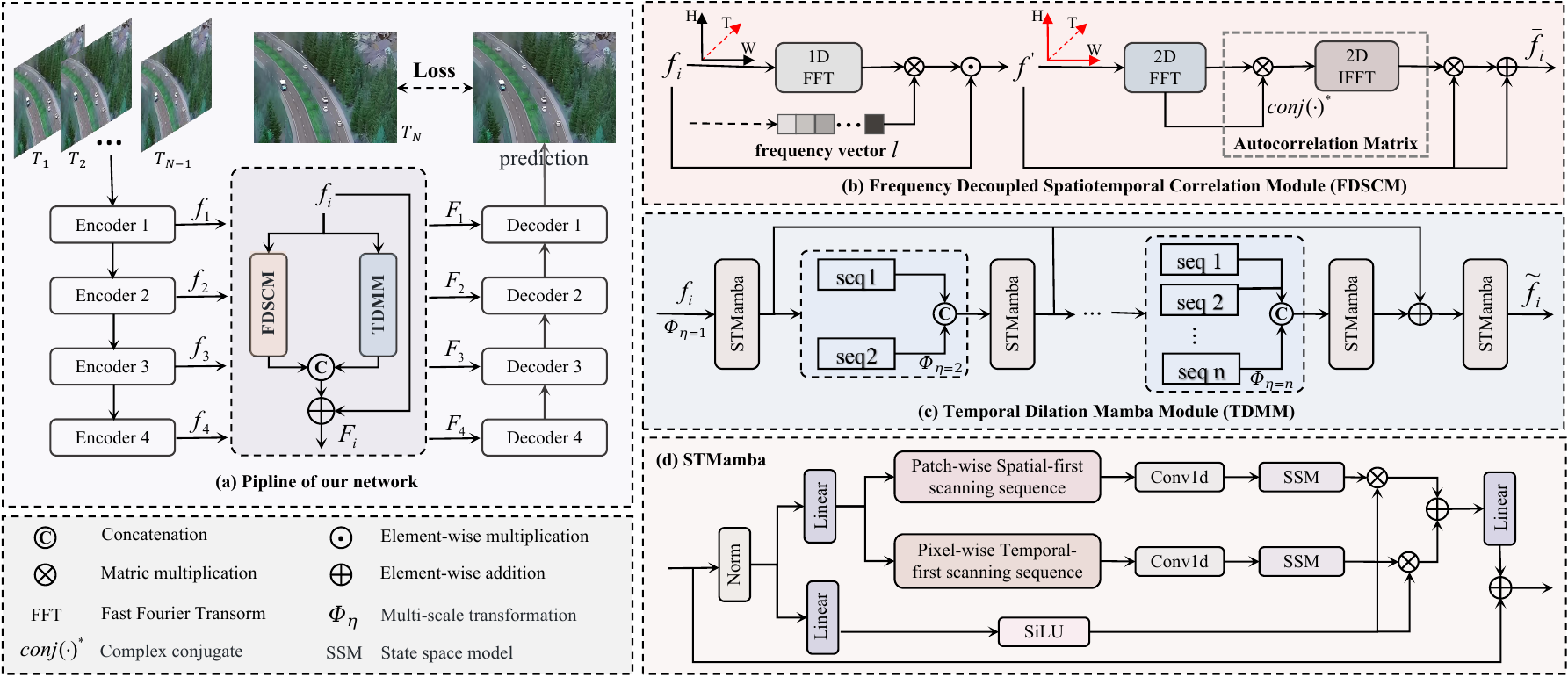} 
        \caption{Overview of the proposed FTDMamba. (a) The overall pipeline adopts an encoder-decoder architecture that contains two core parallel modules: the Frequency Decoupled Spatiotemporal Correlation Module (FDSCM) and the Temporal Dilation Mamba Module (TDMM). (b) The FDSCM first employs a 1D FFT on the temporal dimension for initial multi-source motion decoupling, and subsequently utilizes a 2D FFT on the spatiotemporal domain to model global spatiotemporal dependencies. (c) The TDMM leverages the core STMamba module (d) which constructs distinct pixel-wise temporal-first and patch-wise spatial-first scanning sequences, jointly modeling temporal continuity and local spatial correlations from multiple temporal scales.
        }
	\label{Figure3}
\end{figure*}

\section{Methodology}
\label{sec:Method}
\subsection{Overview}
We treat VAD as a future frame prediction task, aiming to predict the $(N+1)^{th}$ frame from previous $N$ frames. As shown in Fig.  \ref{Figure3}, our proposed framework comprises four parts: Encoder, Decoder, Frequency Decoupled Spatiotemporal Correlation Module (FDSCM) and Temporal Dilation Mamba Module (TDMM). 
The encoder uses a four-stage pyramid transformer~\cite{wang2022pvt} to extract hierarchical features from input frames, which are then processed via two parallel paths. FDSCM leverages frequency domain analysis to decouple multi-source motion components and capture global spatiotemporal dependencies, while TDMM models local spatial correlations and temporal continuity across multiple temporal scales in the spatiotemporal domain. Finally, the complementary outputs are fused and fed to the decoder to predict the future frame. The following subsections detail the design and implementation of the two core modules FDSCM and TDMM.
\subsection{Frequency Decoupled Spatiotemporal Correlation Module}
UAV videos exhibit complex motion coupling, where the UAV's ego-motion and the motion of objects are superimposed at the pixel level, making it challenging to distinguish the sources of motion. Frequency domain analysis provides an effective solution to this challenge, based on the significant differences in spectral structures between different motion patterns. The frequency spectrum of global motion is structured with highly concentrated energy, while the frequency spectrum of local motion is unstructured with dispersed energy. This provides a physically grounded basis for separating coupled motion.
Therefore, we design a Frequency Decoupled Spatiotemporal Correlation Module, comprising temporal frequency decoupling and spatiotemporal correlation modeling.

\subsubsection{Temporal Frequency Decoupling} 
We use FFT along the temporal dimension to decompose aliased motion signals into the frequency domain, and achieve preliminary decoupling of motion sources by adaptively weighting different frequency components.
Given the input feature $f\in \mathbb{R}^{B \times T\times C \times H \times W}$, where $B$ is the batch size, $T$ is the number of frames, $C$ is the number of channels and $H\times W$ is the spatial resolution. We define a normalized frequency vector $l=[l_0, l_1, \cdots, l_{T-1}] \in \mathbb{R}^T$, with elements computed as:
\begin{equation}
	\begin{aligned}
		l_k = \begin{cases}
			\frac{k}{T}, & k = 0, 1, \ldots, \left\lfloor \frac{T}{2} \right\rfloor, \\
			\frac{k - T}{T}, & k = \left\lfloor \frac{T}{2} \right\rfloor + 1, \ldots, T - 1.
		\end{cases}
	\end{aligned}
	\label{eq11}
\end{equation}
Note that $l$ does not depend on the input. For each sample $b$ in the batch, channel c and spatial position ($h$,$w$), we apply the 1D FFT to the temporal sequence $f(b,t,c,h,w)$:
\begin{equation}
	\begin{aligned}
		\hat{f}_k(b,c,h,w) = \sum_{t=0}^{T-1} f(b,t,c,h,w) \cdot e^{-j 2 \pi \frac{k t}{T}},
	\end{aligned}
	\label{eq12}
\end{equation}
where $k=0,1,\ldots,T-1$, $j$ is the imaginary unit, and $\hat{f}_k$ denotes the complex-valued frequency domain feature. Then, we compute the frequency amplitude spectrum:
\begin{equation}
	\begin{aligned}
		A_k(b,c,h,w) = \left| \hat{f}_k \right| = \sqrt{ \operatorname{Re} \big( \hat{f}_k \big)^2 + \operatorname{Im} \big( \hat{f}_k \big)^2 },
	\end{aligned}
	\label{eq13}
\end{equation}
where $Re(\cdot)$ and $Im(\cdot)$  represent the real and imaginary parts, respectively.
To emphasize motion characteristics at different frequencies, we construct frequency-dependent weights $w_k(b,c,h,w) = l_k^2 \cdot A_k(b,c,h,w)^2$, and apply these weights to weight the complex features directly in the frequency domain: $\hat{f}'_k(b,c,h,w) = \hat{f}_k(b,c,h,w) \cdot w_k(b,c,h,w)$.
Then, we apply the inverse FFT (iFFT) to transform the $\hat{f}'_k$ back into the time domain, yielding the weighted feature $f'\in \mathbb{R}^{B \times T\times C \times H \times W}$:
\begin{equation}
	\begin{aligned}
		f'(b,t,c,h,w) = \frac{1}{T} \sum_{k=0}^{T-1} \hat{f}'_k(b,c,h,w) \cdot e^{j 2 \pi \frac{k t}{T}}.
	\end{aligned}
	\label{eq14}
\end{equation}

\subsubsection{Spatiotemporal Correlation Modeling} 
While the aforementioned temporal frequency decoupling can distinguish different types of motion to some extent, it primarily focuses on the temporal evolution characteristics of motion and fails to adequately exploit the spatial structural information. This limitation renders the method ineffective in distinguishing motions that exhibit similar temporal frequency characteristics but differ in spatial distribution patterns. For instance, global high-frequency oscillations induced by UAV jitter and local high-frequency motions of objects in the scene may overlap in the temporal frequency domain, making precise separation unattainable through temporal frequency analysis alone. Therefore, beyond temporal frequency domain analysis, spatiotemporal correlation modeling is needed to fully characterize complex motion patterns.

To model the spatiotemporal correlations of the weighted features $f'$, we leverage the Wiener–Khinchin theorem~\cite{cohen1998generalization} to efficiently compute the autocorrelation matrix in the frequency domain, thereby capturing complex spatiotemporal dependencies. Specifically, we first reshape the spatial dimension into a single dimension $S=H \times W$. For each sample $b$ in the batch and channel $c$, we apply a 2D FFT across both temporal and spatial dimensions of $f'_{(b,c)}(t,s)$:
\begin{equation}
	\begin{aligned}
		\hat{F}'_{(b,c)}(f_t, f_s) = \sum_{t=0}^{T-1} \sum_{s=0}^{S-1} f'_{(b,c)}(t,s) \cdot e^{-j 2 \pi \left( \frac{f_t t}{T} + \frac{f_s s}{S} \right)},
	\end{aligned}
	\label{eq15}
\end{equation}
where $f_t=0,1,\ldots,T-1$ and $f_s=0,1,\ldots,S-1$ denote the temporal and spatial frequency indices, respectively.
Next, we compute the power spectral density (PSD) according to \eqref{eq10}:
\begin{equation}
	\begin{aligned}
		S_{f'}^{(b,c)}(f_t, f_s) = \hat{F}'_{(b,c)}(f_t, f_s) \cdot \hat{F}'_{(b,c)}{}^{*}(f_t, f_s),
	\end{aligned}
	\label{eq16}
\end{equation}
where $\hat{F}'_{(b,c)}{}^{*}$ is the complex conjugate. 
According to the Wiener-Khinchin theorem, we get the spatiotemporal autocorrelation matrix $R_{f'}$ by applying the 2D iFFT to the PSD $S_{f'}$:
\begin{equation}
	\begin{aligned}
		R_{f'} = Re({\frac{1}{TS}}\sum_{t=0}^{T-1} \sum_{s=0}^{S-1} S_{f'}^{(b,c)}(f_t,f_s) \cdot e^{j 2 \pi \left( \frac{f_t t}{T} + \frac{f_s s}{S} \right)}).
	\end{aligned}
	\label{eq17}
\end{equation}
The resulting matrix $R_{f'}$ quantifies the correlation strength between each spatiotemporal position and its neighboring regions. Finally, we use $R_{f'}$ as a spatiotemporal attention map to enhance the feature representation:
$\bar{f} = f' + R_{f'} \odot f'$, where $\odot$ denotes element-wise multiplication.

\begin{figure}[t]
	\centering
	\includegraphics[width=0.48\textwidth,page=1]{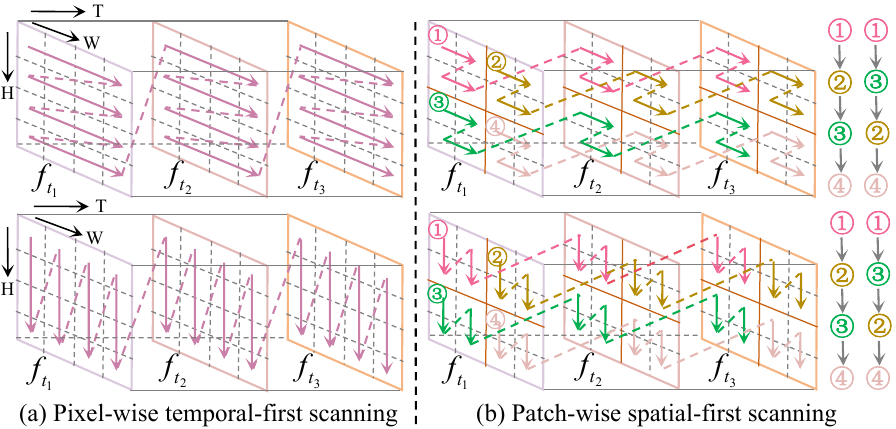} 
	\caption{Illustration of the scanning strategy in STMamba. (a) flattens spatial dimensions in row-major and column-major orders, then concatenates along the temporal axis. (b) divides frame features into patches and extracts temporal sequences at each patch location, then concatenates them in row-major and column-major orders. Here, we only show forward scanning.
	}
	\label{Figure4}
\end{figure}

\subsection{Temporal Dilation Mamba Module}
Although the above FDSCM effectively disentangles motion patterns across different frequency components, frequency domain analysis inherently provides a global representation~\cite{wang2023spatial}, inevitably losing precise spatial location information and motion details while dispersing spatiotemporal information across the entire spectrum. Therefore, to address the limitations of frequency domain analysis, the model requires precise modeling of spatial distributions and temporal dynamic evolution. Leveraging Mamba's~\cite{gu2023mamba} strength in sequence modeling, we propose a Temporal Dilation Mamba Module (TDMM) to efficiently capture inter-frame temporal continuity and local spatial correlations across diverse temporal scales. 

The core of TDMM is the Spatiotemporal Mamba module (STMamba). Given input feature $f \in \mathbb{R}^{B \times T\times C \times H \times W}$, we apply normalization and linear projections to obtain feature representation $X$ and gating feature $Z$. We then construct temporal-first and spatial-first scanning sequences, as illustrasted in Fig. \ref{Figure4}. 
The former flattens the spatial dimensions of each frame in row-major and column-major orders, and concatenate along the temporal axis to create sequences $S_1$, $S_2\in \mathbb{R}^{B \times T \times (H\cdot W)\times C}$. This enables precise tracking of each pixel's temporal evolution and captures fine-grained dynamics.   The latter divides each frame feature into $P \times P$ patches and extracts temporal sequences for each patch location across $T$ frames. The resulting $N = (H/P)\times (W/P)$ patch sequences are concatenated in row-major and column-major orders. Notably, a similar row-major and column-major scanning is also applied to the pixels within each patch, thus producing $S_3$, $S_4$, $S_5$ and $S_6\in \mathbb{R}^{B \times N \times (T\cdot P^2)\times C}$. This design focuses on modeling local spatial correlations and structural consistency.
Each sequence $S_i$ undergoes bidirectional scanning, followed by causal 1D convolution and SSM processing:
\begin{equation}
	\begin{gathered}
		\bar{X}_i = reshape(SSM(Conv1d(S_i))).
	\end{gathered}
	\label{eq18}
\end{equation}
These outputs are gated by $Z$ and summed, then processed by linear mapping and residual connection to get the output:
\begin{equation}
	\begin{gathered}
		out = Linear(\sum_{i=1}^{12} (\sigma(Z) \times \bar{X}_i)) + f,
	\end{gathered}
	\label{eq19}
\end{equation}
where $\sigma$ denotes the $SiLU()$ activation function.

\begin{figure}[t]
	\centering
	\includegraphics[width=0.45\textwidth,page=1]{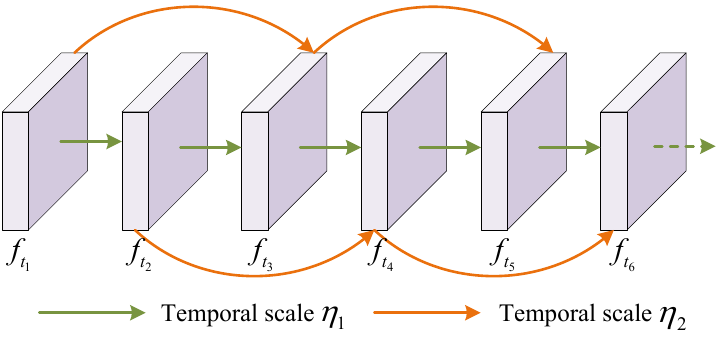} 
	\caption{Multi-scale temporal modeling strategy.  This strategy processes features at different temporal scales by reorganizing consecutive frame features into subsequences with varying temporal resolutions.
	}
	\label{Figure5}
\end{figure}
As discussed earlier, UAV global motion typically presents continuous and smooth patterns, while objects' local motion exhibits discontinuous and irregular patterns. This fundamental difference in temporal characteristics makes single-scale temporal modeling inadequate. To address this issue, we introduce a multi-scale temporal modeling strategy that allows the model to jointly perceive long-term global trends and short-term local variations, as shown in Fig. ~\ref{Figure5}. Specifically, we define a non-parametric reversible transformation $\Phi_{\eta} : \mathbb{R}^{B \times T \times C \times H \times W} \rightarrow \mathbb{R}^{\eta B \times \frac{T}{\eta} \times C \times H \times W}$, where the dilation rate $\eta$ is used to extract frame sequences at different temporal scales. Through $\Phi_{\eta}$, the input features are reorganized into $\eta B$ samples, each forming a subsequence of length $\frac{T}{\eta}$. 
These subsequences are concatenated along the batch dimension and processed independently by a ST-Mamba module, after which an inverse transformation $\Phi_{\eta}^{-1}$ restores their original shape. Finally, features from different temporal scales are fused element-wise and processed through the STMamba module to produce the final output $\tilde{f}$:
\begin{equation}
	\begin{aligned}
		\tilde{f} = \mathrm{STMamba}(\sum_{\eta} \Phi_{\eta}^{-1}(\mathrm{STMamba}(\Phi_{\eta}(f)))).
	\end{aligned}
	\label{eq20}
\end{equation}

\subsection{Decoder}
For multi-scale encoder features $\{f_i\}_{i=1}^4$, we process each through FDSCM and TDMM, then fuse their outputs $\bar{f_i}$ and $\tilde{f_i}$ via channel concatenation and project back to the original dimension, obtaining $\{F_i\}_{i=1}^4$. The fused features are fed to the decoder which consists of hierarchical unConvNormReLU blocks to output the final prediction. Each block incorporates skip connections from the corresponding encoder features $f_i$. The decoding process is defined as:
\begin{equation}
	\begin{aligned}
		\hat{F}_i & = \delta(BN(Up(f_i + F_{i}+ \hat{F}_{i+1}))), i=1, 2, 3, \\
		\hat{F}_4 &= Up(f_4 + F_4), \quad \hat{Y} = Up(Up(\hat{F}_1)),
	\end{aligned}
	\label{eq21}
\end{equation}
where $Up(\cdot)$ is the ConvTranspose2d~\cite{dumoulin2016guide} operation, $BN(\cdot)$ is the batch normalization, $\delta(\cdot)$ is the $Relu()$ activation function, $\hat{Y}$ is the final prediction.

\subsection{Anomaly Inference}
\subsubsection{Loss Functions}
We use intensity loss~\cite{liu2018future} gradient loss $L_{grl}$~\cite{liu2018future}, and structural similarity loss $L_{ssim}$~\cite{lyu2025vadmamba} to make the predicted frame $\hat{Y}$ closer to ground truth $Y$ during training. 
$L_{int}$ uses $l_2$ distance to calculate pixel-wise differences between $\hat{Y}$ and $Y$:
\begin{equation}
	\begin{gathered}
		L_{int}(\hat{Y}, Y) = \|\hat{Y} - Y\|_2^2
		,
	\end{gathered}
	\label{eq22}
\end{equation}
$L_{grl}$ is used to calculate gradient differences between $\hat{Y}$ and $Y$ in both horizontal and vertical directions:
\begin{equation}
\label{eq:23}
\begin{aligned}
L_{grl}(\hat{Y}, Y) &= \sum_{i,j} \left\| | \hat{Y}_{i,j} - \hat{Y}_{i-1,j}| - |Y_{i,j} - Y_{i-1,j} | \right\|_1 \\
&\quad + \left\| | \hat{Y}_{i,j} - \hat{Y}_{i,j-1}| - |Y_{i,j} + Y_{i,j-1}| \right\|_1,
\end{aligned}
\end{equation}
where $i$ and $j$ indicate the spatial index of the frame.
$L_{ssim}$ measures the structural similarity between $\hat{Y}$ and $Y$ at different resolutions. The final loss function can be formulated as the weighted sum of the above three losses:
\begin{equation}
	\begin{gathered}
		L = \alpha L_{int}(\hat{Y}, Y) + \beta L_{grl}(\hat{Y}, Y) + \gamma L_{ssim}(\hat{Y}, Y),
	\end{gathered}
	\label{eq24}
\end{equation}
where $\alpha$, $\beta$ and $\gamma$ are loss weighting parameters.

\subsubsection{Anomaly Scores}
During the inference stage, we employ the difference between predicted frame $\hat{Y}$ and ground truth $Y$ to detect anomalies. Following the works~\cite{le2025hstforu, cai2021appearance}, we adopt Peak Signal to Noise Ratio (PSNR) to assess the image quality:
\begin{equation}
	\begin{gathered}
		\text{PSNR}(Y, \hat{Y}) = 10 \log_{10} \frac{\left[\max _{\hat{Y}}\right]^2}{\frac{1}{N} \sum_{i=0}^{N-1} \left(Y_i - \hat{Y}_i\right)^2},
	\end{gathered}
	\label{eq25}
\end{equation}
where N is the number of pixels in the frame, $\max_{\hat{Y}}$ is the maximum pixel value in $\hat{Y}$.
A higher PSNR represents the frame is more likely to be normal. 
After calculating the PSNR values for each testing frame, we normalize them to the range [0,1] and calculate the normal score for each frame:
\begin{equation}
	\begin{gathered}
		S(t) = \frac{\text{PSNR}_t - \text{PSNR}_{min}}{\text{PSNR}_{max} - \text{PSNR}_{min}},
	\end{gathered}
	\label{eq26}
\end{equation}
where $\text{PSNR}_t$ is the PSNR value of $t^{th}$ frame, $\text{PSNR}_{min}$ and $\text{PSNR}_{max}$ denote the minimum and the maximum PSNR values in the given video sequence.
Based on the calculated score for each frame, we can set a threshold to distinguish  abnormal frames.

\section{MUVAD Dataset}
\label{sec1}

Existing UAV VAD datasets, such as Drone-Anomaly~\cite{jin2022anomaly} and UIT-ADrone~\cite{tran2023uit}, are mainly captured by hovering UAVs, limiting their applicability in real-world scenarios. In practical deployments, UAVs often require continuous motion for tasks. This motion pattern introduces dynamic backgrounds, posing challenges to existing VAD methods designed for static backgrounds. To bridge the gap between academic researches and practical applications, we introduce MUVAD, a large-scale benchmark designed for dynamic aerial surveillance.

MUVAD primarily focuses on anomaly detection within the traffic domain, which is a critical aspect of urban safety and public administration. In this domain, anomalous events occur with high frequency and are clearly defined, allowing for data collection through both controlled filming and publicly available sources. In contrast, other application areas such as environmental monitoring, industrial inspection, or maritime security often experience low anomaly incidence, making data collection challenging. Additionally, these fields are frequently subject to stringent privacy and security regulations that limit data accessibility. Therefore, traffic anomaly detection serves as a suitable choice for dynamic UAV VAD research. Next, we will introduce the details of MUVAD.

\begin{figure}[t]
	\centering
	\includegraphics[width=0.48\textwidth,page=1]{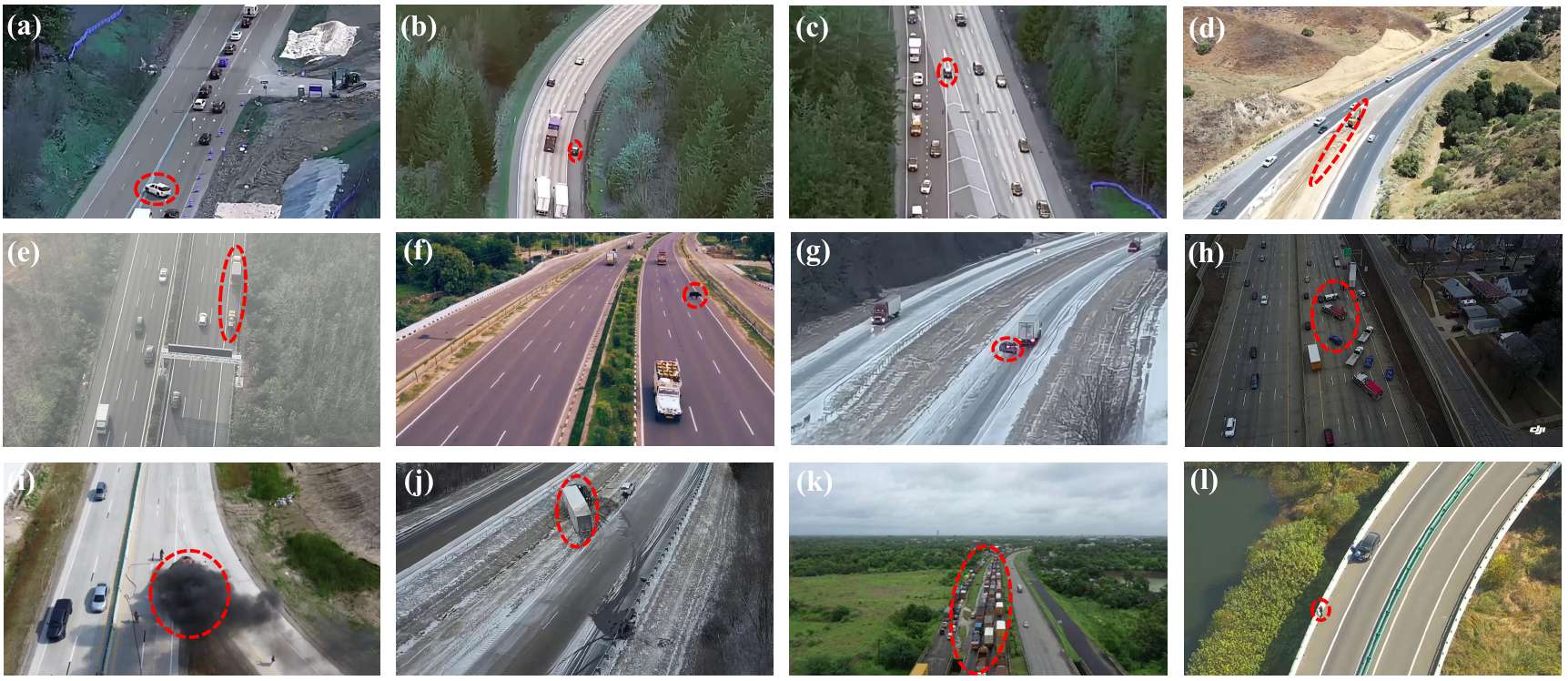} 
	\caption{Visualization of anomalies in the MUVAD Dataset, including 12 anomaly categories: (a) Illegal Lane Change, (b) Emergency Lane Violation, (c) Wrong-way Driving, (d) Construction Zone, (e) Vehicle Breakdown, (f) Animal Intrusion, (g) Vehicle Skidding, (h) Vehicle Collision, (i) Fire Incident, (j) Roadside Deviation,   (k) Traffic Congestion, and (l) Pedestrian Intrusion.  
	}
	\label{Figure6}
\end{figure}

\begin{table}[t]
    \caption{
Statistics of Anomaly Event Types and the Corresponding Event Numbers in the MUVAD Dataset.
		}
        \setlength{\tabcolsep}{8.0pt}
	\begin{center}

		\begin{tabular}{c c c }
			\toprule[1pt]
			Index & Anomaly Event Type         & Number of Events \\ 
                \midrule
			1     & Illegal Lane Change        & 21  \\ 
			2     & Emergency Lane Violation  & 39  \\ 
			3     & Wrong-way Driving          & 15  \\
                4     & Construction Zone         & 47   \\
                5     & Vehicle Breakdown         & 9   \\
                6     & Animal Intrusion           & 6   \\
                7     & Vehicle Skidding           & 16   \\
                8     & Vehicle Collision                 & 12   \\
                9     & Fire Incident                      & 7   \\
                10    & Roadside Deviation         & 11   \\
                11    & Traffic Congestion               & 16   \\
                12    & Pedestrian Intrusion   & 41   \\
            \bottomrule[1pt]
		\end{tabular}
		
		\label{tab:table1}
	\end{center}
\end{table}

\subsection{Data Collection and Annotation}
The video data in MUVAD originates from two main channels: UAV-filmed real-world scenarios, including urban roads, transportation hubs, and vehicle-dense areas, and publicly available videos from platforms like YouTube to enrich anomalous event variety. To ensure dataset quality, we exclude videos with insufficient duration, blurriness, heavy edits, or non-UAV footage.
Following standard unsupervised learning protocols in VAD~\cite{jin2022anomaly,tran2023uit,lu2013abnormal,luo2017revisit}, the training set contains only normal samples, while the test set includes both normal and anomalous frames, labeled '0' for normal and '1' for anomalies. We employ a multi-annotator cross-validation to ensure annotation accuracy, with disputed samples reviewed by experts to finalize the labels.

\subsection{Dataset Statistics}

MUVAD comprises 222,736 video frames in total, with the training set containing 46 video clips (126,254 frames) and the testing set containing 72 video sequences (96,482 frames). Each video has a frame rate of 30 frames per second and a spatial resolution of $852 \times 480$ pixels. The dataset encompasses diverse complex scenarios including urban arterial roads, transportation hubs, high-traffic areas, and construction zones. These scenes were captured under diverse conditions of time (day/night), weather (sunny/foggy), and UAV motion patterns (translation/rotation/tracking). As illustrated in Fig. \ref{Figure6} and Table \ref{tab:table1}, MUVAD defines 12 typical anomaly categories with 240 anomaly event instances, including illegal lane change, emergency lane violation, wrong-way driving, construction zone, vehicle breakdown, animal intrusion, vehicle skidding, vehicle collision, fire incident, roadside deviation, traffic congestion, and pedestrian intrusion.

\subsection{Comparison with Related Datasets}
As detailed in Table \ref{tab:table2}, we conduct a comparative analysis of MUVAD against four representative VAD datasets, including two ground-based surveillance datasets  and two UAV VAD datasets.
In terms of scale, MUVAD presents advantages over both Drone-Anomaly and UIT-ADrone. A more critical distinction lies in the background dynamics. The CHUK Avenue, ShanghaiTech, and UIT-ADrone datasets exclusively feature static backgrounds. Although Drone-Anomaly incorporates some footage from moving UAVs, its overall scale is limited and dynamic background samples constitute only $35.4\%$ of total frames. 
Furthermore, MUVAD defines 12 anomaly types with 240 events. While the total number of events is less than that in UIT-ADrone, the anomalies in MUVAD present greater detection challenges and hold higher practical value, as they are embedded within complex dynamic backgrounds.

\section{Experiments}
\subsection{Experimental Setup}
\subsubsection{Datasets}
To  evaluate the performance of FTDMamba, we conduct experiments on three UAV VAD datasets, including the publicly available Drone-Anomaly and UIT-ADrone datasets, as well as our newly constructed MUVAD dataset. 
\begin{itemize}
    \item \textbf{Drone-Anomaly dataset} comprises 37 training and 22 testing video sequences, containing 51,635 training and 35,853 testing frames at a resolution of 640×640 pixels.  It covers ten categories of anomalous events, such as panel defects, railway obstacles, and unidentified objects. Notably, 18 training and 9 testing video sequences contain dynamic backgrounds, accounting for approximately $35.4\%$ of the total video frames.
    \item \textbf{UIT-ADrone dataset} consists of 41 training and 51 testing video sequences, totaling 59,186 training frames and 147,005 testing frames at a resolution of 1920×1080 pixels. The dataset covers 10 types of anomalous events such as wrong-way driving in roundabouts, carrying bulky goods, and falling off motorcycles. All video sequences are captured by stationary hovering UAVs.
    \item Unlike the two datasets mentioned above, all videos in the \textbf{MUVAD dataset} are captured by continuously moving drones, providing a challenging benchmark for evaluating UAV VAD methods under dynamic scenarios. Detailed information about MUVAD is presented in Section \ref{sec1}.
    
\end{itemize}

\begin{table*}[t]
    \renewcommand{\arraystretch}{0.95}
    \caption{
            Comparison of the new MUVAD benchmark with other VAD datasets.
        }
    \begin{center}
    \renewcommand\tabcolsep{5pt}
        
        \small
        \begin{tabular}{c|c|c|c|c|c|c|c|c|c}
        
            \toprule[1pt]
            \multirow{2}{*}{Dataset}  & Video Snippets & Total & Training & Testing & Anomaly & Anomaly  & \multirow{2}{*}{Source} & \multirow{2}{*}{Background} & \multirow{2}{*}{Year} \\
& (Train / Test) & Frames & Frames & Frames & Types & Events & & & \\ 
            \midrule
            
            CHUK Avenue~\cite{lu2013abnormal}    & 16 / 21   & 30,652  & 15,328  & 15,324  & 5   & 77   & Ground & Static  & 2015 \\ 
            ShanghaiTech~\cite{luo2017revisit}  & 238 / 199         & 317,398 & 274,515 & 42,883  & 11  & 158  & Ground & Static  & 2017 \\ 
            Drone-Anomaly~\cite{jin2022anomaly} & 37 / 22   & 87,488  & 51,635  & 35,853  & 10   & 26    & Drone  & Mixed   & 2022 \\ 
            UIT-ADrone~\cite{tran2023uit}    & 41 / 51   & 206,194 & 59,186  & 147,005 & 10  & 1,935 & Drone  & Static  & 2023 \\ 
            MUVAD (ours)  & 46 / 72   & 222,736 & 126,254 & 96,482  & 12  & 240  & Drone  & Dynamic & 2025 \\ 
            \bottomrule[1pt]
        \end{tabular}
        
         \label{tab:table2}
    \end{center}
   
\end{table*}

\begin{table*}[t]
    \renewcommand{\arraystretch}{1.07}
    \renewcommand\tabcolsep{6.5pt}
    \small
    \begin{center}
      
        \caption{Quantitative comparison of our proposed FTDMamba with existing VAD methods on the three benchmarks. Competing methods include those designed for ground surveillance and UAV-specific scenarios. Performance is evaluated using Micro-AUC ($\%$), Macro-AUC ($\%$), and EER. The best results are highlighted in \textbf{bold}, and the second-best results are \underline{underlined}.}
            \begin{tabular}{c|c|c|ccc|ccc|ccc}
                \toprule[1pt]
                & \multirow{2}[2]{*}{Method} & \multirow{2}[2]{*}{Year} & \multicolumn{3}{c|}{Drone-Anomaly}          & \multicolumn{3}{c|}{UIT-ADrone}        & \multicolumn{3}{c}{MUVAD}   \\
                \cmidrule{4-6} \cmidrule{7-9} \cmidrule{10-12} & & & Micro$\uparrow$ & Macro$\uparrow$ & EER$\downarrow$ & Micro$\uparrow$ & Macro$\uparrow$ &EER$\downarrow$ & Micro$\uparrow$ & Macro$\uparrow$ & EER$\downarrow$ \\
                \midrule
                \multirow{9}{*}{\rotatebox{90}{Ground Surveillance VAD}} 
                & AMMCNet~\cite{cai2021appearance}      &2021 & 62.4 & 63.6 & 0.457 & 60.7  & 59.2 & 0.441  & 57.2 & 55.6 & 0.583  \\
                & STD~\cite{chang2022video}             &2022 & 58.5 & 57.4 & 0.512 & 59.1  & 58.6 & 0.459  & 60.4 & 61.2 & 0.506   \\
                & F$_2$DN~\cite{9622181}                &2022 & 56.0 & 59.7 & 0.495 & 59.6  & 60.3 & 0.448  & 61.4 & 60.2 & 0.490 
                \\
                & MAAMNet~\cite{wang2023memory}         &2023 & 61.5 & 60.4 & 0.473 & 60.1  & 61.8 & 0.462  & 62.8 & 60.5 & 0.524  \\
                & ASTNet~\cite{le2023attention}         &2023 & 63.2 & 64.9 & 0.422 & 61.4  & 63.7 & 0.415  & 60.3 & 63.4 & 0.487  \\
                & VAD-Cluster~\cite{qiu2024video}       &2024 & 63.6 & 60.8 & 0.432 & 58.4  & 57.6 & 0.494  & 59.5 & 61.7 & 0.539  \\
                & VAD-Mamba~\cite{lyu2025vadmamba}      &2025 & 66.7 & 68.1 & 0.393 & \underline{68.6}  & 66.5 & 0.403  & 66.2 & 62.6 & 0.437 
                \\
                & LGN-NET~\cite{zhao2025rethinking}     &2025 & 64.3 & 63.2 & 0.455 & 62.7  & 64.4 & 0.428  & 63.5 & 64.8 & 0.463  \\
                & MA-PDM~\cite{zhou2025video}           &2025 & \underline{68.3} & 67.5 & \underline{0.360} & 65.7  & 67.2 & 0.384  & 65.6 & 63.4 & 0.432 
                \\
                \midrule
                \multirow{4}{*}{\rotatebox{90}{UAV VAD}} 
                & ANDT~\cite{jin2022anomaly}            &2022 & 67.4 & 66.9 & 0.452 & 60.5 & 63.2 & 0.424   & 65.4 & 63.3 & 0.455  \\
                & ASTT~\cite{tran2024transformer}       &2024 & 67.8 & 68.7 & 0.397 & 65.5 & 66.0 & 0.392   & \underline{67.9} & \underline{65.1} & \underline{0.413}  \\
                & HSTforU~\cite{le2025hstforu}          &2025 & 68.0 & \underline{69.2} & 0.372 & 66.2 & \underline{68.2} & \underline{0.386}   & 66.1 & 64.7 & 0.429  \\
                & Ours     & - - & \textbf{71.6} & \textbf{72.3} & \textbf{0.336} & \textbf{70.7} & \textbf{69.5} & \textbf{0.368}  & \textbf{71.4} & \textbf{68.4} & \textbf{0.372}  \\ 
                \bottomrule[1pt]
            \end{tabular}
        \label{tab:table3}
    \end{center}
\end{table*}

\subsubsection{Evaluation Metric}
We employ the Area Under the Receiver Operating Characteristic Curve (AUC) as the primary evaluation metric, where a higher value indicates better performance.
Following  prior work~\cite{cao2024context,ristea2024self}, we report both Micro-AUC and Macro-AUC. Micro-AUC computes a single global score by aggregating frames from all testing videos, whereas Macro-AUC first computes an AUC for each testing video and then averages these values.
Furthermore, we also report the Equal Error Rate (EER)~\cite{tran2024transformer}, defined as the error rate at which the False Positive Rate equals the False Negative Rate. A lower EER reflects better prediction results. For ablation studies, we report only the Micro-AUC.

\subsubsection{Implementation Details}
Our model is implemented in the PyTorch deep learning framework and trained on two NVIDIA GeForce RTX 3090 GPUs. During data preprocessing, all video frames are resized to 256×256 pixels and normalized to the range [-1, 1]. Following the future frame prediction paradigm, the model takes six consecutive frames as input to predict the seventh frame.
For training configuration, we set the batch size to 8 and use the AdamW optimizer~\cite{yao2021adahessian} with a cosine annealing learning rate schedule~\cite{loshchilov2017sgdr} over 200 epochs. The initial learning rate is set to 0.00005 for Drone-Anomaly and MUVAD, and 0.0001 for UIT-ADrone.
In the TDMM module, the depth of submodule STMamba is set to 1. For patch-wise spatial-first scanning strategy, the patch size is set to 4 to capture local spatial structures. The multi-scale temporal modeling employs dilation rates $\eta = \{1, 2, 3\}$.

\subsection{Quantitative Comparison with Existing Methods}
To assess FTDMamba’s performance, we conduct quantitative comparisons against recently proposed VAD methods.
The compared methods fall into two major categories: (1) advanced approaches originally designed for ground-based surveillance videos, including AMMCNet~\cite{cai2021appearance}, STD~\cite{chang2022video}, F$_2$DN~\cite{9622181}, MAAMNet~\cite{wang2023memory},  ASTNet~\cite{le2023attention}, VAD-Cluster~\cite{qiu2024video}, VAD-Mamba~\cite{lyu2025vadmamba}, LGN-Net~\cite{zhao2025rethinking}, and  MA-PDM~\cite{zhou2025video}; and (2) the recent methods developed for UAV videos, such as  ANDT~\cite{jin2022anomaly}, ASTT~\cite{tran2024transformer}, and HSTforU~\cite{le2025hstforu}.
To ensure fair comparison, we adopt the published evaluation metric results from the original papers where available. Otherwise, we retrain the models using their official codes and parameter settings.

As shown in Table \ref{tab:table3}, FTDMamba achieves SOTA performance across three datasets. On the Drone-Anomaly dataset, our method attains Micro-AUC of $71.6\%$  and Macro-AUC of $72.3\%$,  surpassing the UAV-based method HSTforU by $5.3\%$ and $4.5\%$, and outperforming the ground-based method MA-PDM by $4.8\%$ and $7.1\%$, respectively. On the UIT-ADrone dataset, our method surpasses VAD-Mamba by $3.1\%$ in Micro-AUC and HSTforU by $1.9\%$ in Macro-AUC. On the MUVAD dataset with dynamic backgrounds, our method reaches $71.4\%$ and $68.4\%$ in Micro-AUC and Macro-AUC, exceeding the second-best method ASTT by $5.2\%$ and $5.1\%$. Notably, in terms of the metric EER, FTDMamba yields the lowest EER among all compared methods on three datasets.
Overall, these results demonstrate that FTDMamba performs robustly under both dynamic and static conditions, establishing it as a generalizable and efficient solution for UAV VAD. 
To further evaluate the robustness of FTDMamba, we compare performance of different methods across four scenes from the Drone-Anomaly dataset, with results presented in Table \ref{tab:table4}. These scenes encompass Highway, Bike Roundabout, Vehicle Roundabout, and Crossroads.
Across the first three scenarios, FTDMamba achieves superior performance compared to competing methods. The Crossroads scenario presents the most significant challenge, where complex vehicle trajectories and interactions lead to performance degradation across all methods. Nevertheless, FTDMamba maintains the highest Macro-AUC of $62.9\%$ and achieves a competitive EER of 0.403, demonstrating more balanced and stable anomaly detection capability in complex environments.

\begin{table*}[t]
	\renewcommand{\arraystretch}{1.10}
    \renewcommand\tabcolsep{5pt}
	\begin{center}
		\small
            \caption{Scene-specific performance comparison of different VAD methods on the Drone-Anomaly dataset, covering four distinct scenarios: Highway, Bike roundabout, Vehicle roundabout, and Crossroads.}
		\begin{tabular}{c|c|ccc|ccc|ccc|ccc}
			\toprule[1pt]
			& \multirow{2}[2]{*}{Method}           & \multicolumn{3}{c|}{Highway}        & \multicolumn{3}{c|}{Bike Roundabout} & \multicolumn{3}{c|}{Vehicle Roundabout} & \multicolumn{3}{c}{Crossroads}   \\
            \cmidrule{3-5} \cmidrule{6-8} \cmidrule{9-11} \cmidrule{12-14} 
            & & Micro & Macro & EER & Micro & Macro & EER & Micro & Macro & EER & Micro & Macro & EER \\ 
			\midrule
			\multirow{9}{*}{\rotatebox{90}{Ground Surveillance VAD}} 
			& AMMCNet~\cite{cai2021appearance} & 66.5 & 68.8 & 0.313 & 64.1 & 65.6 & 0.448 & 63.9 & 65.5 & 0.383 & 42.6 & 43.4 & 0.517   \\
                & STD~\cite{chang2022video}        & 76.0 & 74.7 & 0.295 & 79.5 & 76.4 & 0.232 & 38.6 & 37.7 & 0.617 & 38.1 & 39.7 & 0.594   \\
                & F$_2$DN~\cite{9622181}         & 62.8 & 63.4 & 0.369 & 64.0 & 67.5 & 0.406 & 61.4 & 60.1 & 0.411 & 44.8 & 51.3 & 0.540   \\
                & MAAMNet~\cite{wang2023memory}           & 70.3 & 68.5 & 0.323 & 71.5 & 69.8 & 0.326 & 56.7 & 58.2 & 0.465 & \underline{58.4} & 55.7 & 0.412  \\
                & ASTNet~\cite{le2023attention}            & 65.2 & 70.3 & 0.357 & 74.7 & 71.4 & 0.274 & 62.3 & 66.4 & 0.378 & 54.4 & 56.2 & 0.432   \\
                & VAD-Cluster~\cite{qiu2024video}       & 67.9 & 61.2 & 0.372 & 69.8 & 66.4 & 0.381 & 64.8 & 68.5 & 0.342 & 49.6 & 52.4 & 0.435   \\
			& VAD-Mamba~\cite{lyu2025vadmamba}        & 77.5 & \underline{79.4} & 0.275 & 76.3 & 78.7 & 0.265 & 67.2 & 71.6 & 0.301 & 52.7 & 53.5 & 0.462  \\
                & LGN-NET~\cite{zhao2025rethinking}          & 67.4 & 70.3 & 0.304 & 69.7 & 71.5 & 0.362 & 66.3 & 68.6 & 0.356 & 50.3 & 48.6 & 0.475  \\
                & MA-PDM~\cite{zhou2025video}           & 76.8 & 78.3 & \underline{0.265} & 80.4 & {78.6} & 0.249 & 70.6 & 67.9 & 0.314 & 56.2 & 58.6 & 0.423  \\
			\midrule
			\multirow{4}{*}{\rotatebox{90}{UAV VAD}} 
                & ANDT~\cite{jin2022anomaly}            & 68.7 & 70.2 & 0.315   & 82.2 & 79.8 & 0.237 & 61.3 & 64.3 & 0.378 & \textbf{65.2} & \underline{61.3} & \textbf{0.376}  \\
			& ASTT~\cite{tran2024transformer}             & \underline{83.5} & 79.1 & 0.274 & 78.4 & \underline{81.2} & 0.252 & \underline{74.1} & \underline{75.6} & \underline{0.285} & 38.6 & 36.4 & 0.562  \\
                & HSTforU~\cite{le2025hstforu}        & 70.6 & 72.5 & 0.286 & \underline{82.7} & {80.4} & \textbf{0.226} & 65.6 & 69.7 & 0.322 & 53.5 & 54.3 & 0.467   \\
			& Ours     & \textbf{84.8} & \textbf{85.3} & \textbf{0.258} & \textbf{84.8} & \textbf{83.2} & \underline{0.235} & \textbf{77.8} & \textbf{78.2} & \textbf{0.276} & {57.0} & \textbf{62.9} & \underline{0.403}\\ 
			\bottomrule[1pt]
		\end{tabular}
		
		\label{tab:table4}
	\end{center}
\end{table*}

\subsection{Qualitative Results Comparison and Analysis} 
\begin{figure*}[t]
	\centering
	\includegraphics[width=0.9\textwidth,page=1]{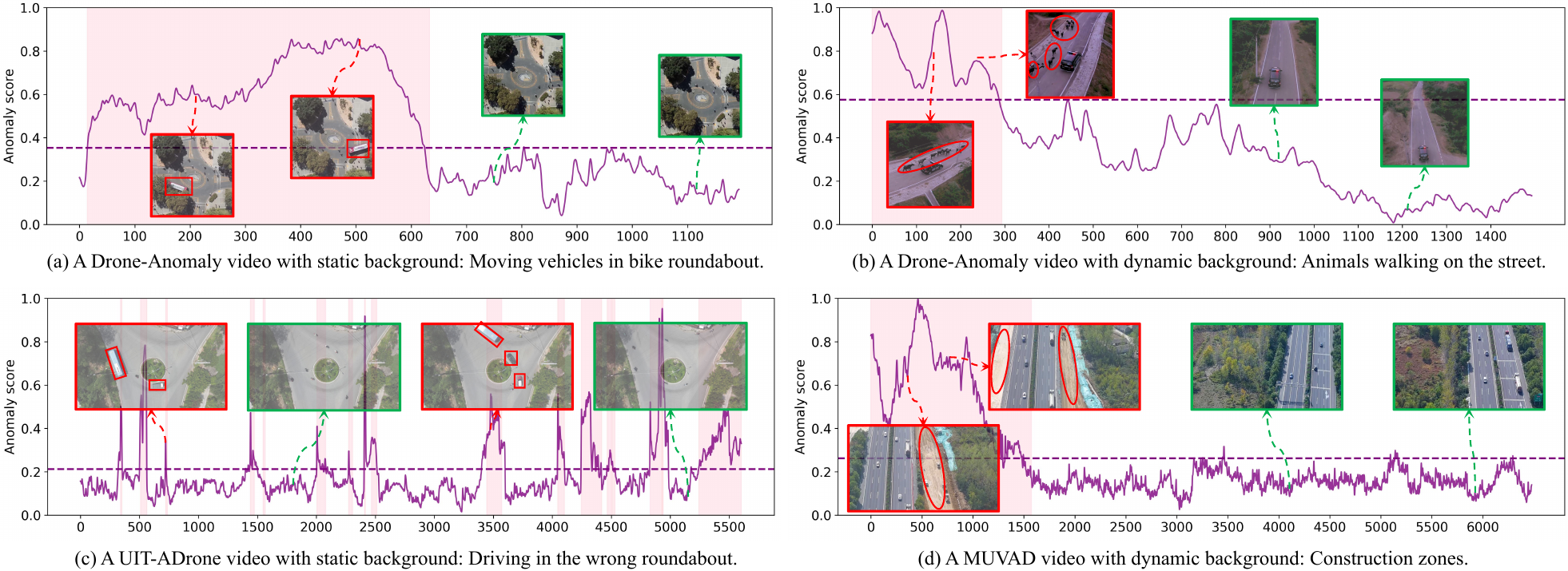} 
	\caption{The anomaly score curves of some testing videos on the Drone-Anomaly, MUVAD and UIT-ADrone datasets. The red regions represent the anomalous intervals, and the purple dashed line indicates the adaptive threshold.}
	\label{Figure7}
\end{figure*}

In this section, we demonstrate the effectiveness of FTDMamba through visualization.  
First, we analyze the anomaly curves of FTDMamba on four testing videos to evaluate temporal localization precision. 
We employ an adaptive threshold~\cite{lyu2025vadmamba} to dynamically determine anomaly criteria. Since anomaly scores are normalized to [0, 1], we generate 100 uniformly distributed thresholds spanning this interval. For each threshold, we derive predicted labels from the anomaly scores and compute their F1 scores against ground truth labels. The threshold yielding the highest F1 score is optimal.
As shown in Fig. \ref{Figure7}, the model achieves high precision under static background conditions. In Fig. \ref{Figure7}(a), for the anomaly of moving vehicles in bike roundabout, the anomaly score curve remains consistently below the threshold during normal periods and rises sharply upon anomaly occurrence. In Fig. \ref{Figure7}(c), the model successfully detects multiple discrete instances of vehicles driving incorrectly through a roundabout.
In more challenging dynamic backgrounds, the model still exhibits strong robustness. Fig. \ref{Figure7}(b) presents a scenario captured by a moving UAV. Despite the fluctuations introduced by the UAV's ego-motion,  FTDMamba produces anomaly scores exceeding the threshold when animals appear, validating the method's ability to distinguish background motion from true anomalous behavior.  Furthermore,  Fig. \ref{Figure7}(d) shows that when confronted with a prolonged construction zone anomaly, the model responds promptly at onset and maintains consistently elevated scores throughout the anomaly duration, demonstrating sustained recognition capability for long-duration anomalies. 

\begin{figure*}[t]
	\centering
	\includegraphics[width=0.9\textwidth,page=1]{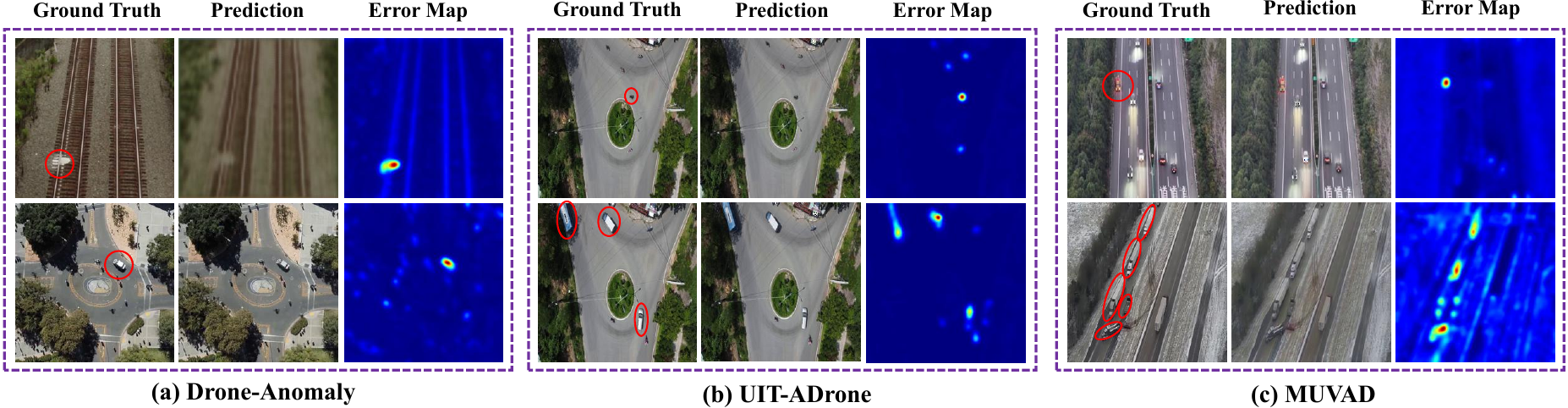} 
	\caption{Visualization of the predicted frames and their corresponding prediction error maps on the Drone-Anomaly, UIT-ADrone, and MUVAD datasets. Anomalous regions are highlighted by red circles.
	} 
	\label{Figure8}
\end{figure*}

We further demonstrate the model's spatial anomaly localization capability through predicted frames, ground truth, and error maps.  The error maps are generated by computing the pixel-wise differences between ground-truth and predicted frames, where high-intensity regions indicate larger prediction errors. As illustrated in Fig. \ref{Figure8}, for normal regions, the error maps appear predominantly blue, indicating minimal pixel-level discrepancies between predictions and ground truth. In contrast, the model fails to accurately predict anomalous regions, which are visualized as high-intensity regions in the error maps. For instance, in the second row of Fig. \ref{Figure8}(b), the model fails to correctly predict the moving vehicles in bike roundabout, producing highlighted regions in the error map at their locations.

\subsection{Ablation Study}
\subsubsection{Effectiveness of FDSCM}

\begin{figure}[t]
	\centering
	\includegraphics[width=0.45\textwidth,page=1]{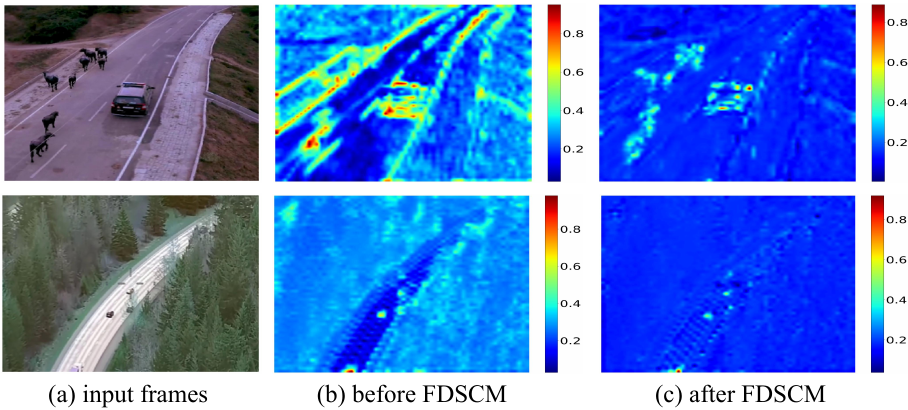} 
	\caption{Visualization of the FDSCM in motion decoupling. (a) shows the input frames, (b) shows the feature maps before entering FDSCM, (c) shows the feature maps after processing by FDSCM.
	}      
	\label{Figure9}
\end{figure}

We conduct ablation studies to verify the effectiveness of FDSCM and the contribution of its two core components: Temporal Frequency Decoupling (TFD) and Spatio-Temporal Correlation modeling (STC).
As shown in Table \ref{tab:table5}, integrating FDSCM into the baseline (Task A→B) improves Micro-AUC by $3.1\%$ on UIT-ADrone and Macro-AUC by $5.8\%$ on MUVAD. 
To further disentangle the source of these gains, we individually remove TFD and STC from the complete model (Task G). Removing TFD (Task G→E) decreases performance by $2.6\%$ and $4.2\%$ on the two datasets, respectively, while removing STC (Task G→F) reduces by $2.3\%$ and $3.3\%$. 
Fig. \ref{Figure9} visualizes FDSCM's effectiveness in decoupling motion features. As depicted in Fig. \ref{Figure9}(b), without FDSCM processing, the global background flow induced by UAV ego-motion produces large-scale, diffuse activations in the original feature maps. After processed by FDSCM, background activations are substantially suppressed, focusing attention on foreground moving objects.

\begin{table}[t]
    \setlength{\tabcolsep}{5pt}
    \caption{
            Comparison of each module's performance on UIT-ADrone and MUVAD. TFD indicates the Temporal Frequency Decoupling, STC indicates the Spatiotemporal Correlation modeling, STM indicates the STMamba, and MST indicates the Multi-scale Temporal Modeling.
        }
    \begin{center}
        \small
        \renewcommand{\arraystretch}{0.9}
        \begin{tabular}{ c| c c c c |c c}
            \toprule[1pt]
            \multirow{2}[2]{*}{Task} & \multicolumn{2}{c}{FDSCM} & \multicolumn{2}{c|}{TDMM} & \multirow{2}[2]{*}{UIT-ADrone} & \multirow{2}[2]{*}{MUVAD} \\
            \cmidrule{2-3} \cmidrule{4-5}
            & TFD & STC & STM & MST &  &  \\
            \midrule
            A & \ding{55} & \ding{55} & \ding{55} & \ding{55} & 61.5 & 60.7 \\
            B & \ding{51} & \ding{51} & \ding{55} & \ding{55} & 63.4 & 64.2 \\
            C & \ding{51} & \ding{51} & \ding{51} & \ding{55} & 67.1 & 68.8 \\
            D & \ding{55} & \ding{55} & \ding{51} & \ding{51} & 67.6 & 66.9 \\
            E & \ding{55} & \ding{51} & \ding{51} & \ding{51} & 68.9 & 68.5 \\
            F & \ding{51} & \ding{55} & \ding{51} & \ding{51} & 69.1 & 69.1 \\
            G & \ding{51} & \ding{51} & \ding{51} & \ding{51} & \textbf{70.7} & \textbf{71.4} \\
            \midrule
            H & \multicolumn{4}{c|}{Cascaded Structure} & 69.3 & 68.5 \\
            \bottomrule[1pt]
            
        \end{tabular}
        
        \label{tab:table5}
    \end{center}
\end{table}

\begin{table}[t]
    \caption{The impact of different scanning strategies. TFScan indicates temporal-first scanning sequences, and SFScan indicates spatial-first scanning sequences.}
	\setlength{\tabcolsep}{6pt}
	\small
        \renewcommand{\arraystretch}{0.9}
	\begin{center}
		\begin{tabular}{c| cc| cc}
        
			\toprule[1pt]
			\multirow{2}[2]{*}{Task} & \multicolumn{2}{c|}{Scanning Strategies} & \multirow{2}[2]{*}{UIT-ADrone} & \multirow{2}[2]{*}{MUVAD} \\
			\cmidrule(lr){2-3}
			& TFScan           & SFScan           &                      &                      \\
			\midrule
			A                      & pixel-wise        & patch-wise        & \textbf{70.7}                 & \textbf{71.4}                 \\
			B                      & patch-wise        & pixel-wise        & 69.5                 & 70.2                 \\
			C                      & pixel-wise        & pixel-wise        & 70.2                 & 70.8                 \\
			D                      & patch-wise        & patch-wise        & 68.8                 & 68.6                 \\
			\bottomrule[1pt]
		\end{tabular}
		
		\label{tab:table6}
	\end{center}
\end{table}

\begin{table}[t]
    \caption{
			The impact of STMamba's depth on performance.
		}
        \setlength{\tabcolsep}{7.0pt}
	\begin{center}
    \renewcommand{\arraystretch}{0.9}

		\small
		\begin{tabular}{c| c| c c| c}
			\toprule[1pt]
			Task & Depth       & UIT-ADrone & MUVAD  & FPS \\ \midrule
			A     & [1,1,1,1] & 70.7       & 71.4  & \textbf{29}  \\ 
			B     & [2,2,2,2] & 71.4          & 70.9     & 21  \\ 
			C     & [3,3,3,3] & \textbf{71.9}          & \textbf{71.7}     & 11   \\ \bottomrule[1pt]
		\end{tabular}
		
		\label{tab:table7}
	\end{center}
\end{table}

\begin{table}[t]
	\renewcommand{\arraystretch}{0.9}
    \caption{
			The impact of different combination of loss functions.
		}
    \setlength{\tabcolsep}{7pt}
	\begin{center}

		\small
		\begin{tabular}{c| ccc| c c}
			\toprule[1pt]
			\multirow{2}[2]{*}{Task} & \multicolumn{3}{c|}{Loss Function} & \multirow{2}[2]{*}{UIT-ADrone} & \multirow{2}[2]{*}{MUVAD} \\ \cmidrule(lr){2-4}
			& $L_{int}$ & $L_{grl}$ & $L_{ssim}$ &           &    \\ \midrule    
			A     & \ding{51}  & \ding{55}     & \ding{55}      &  68.5          & 69.3     \\ 
			B     & \ding{51}  & \ding{51}  & \ding{55}       &  70.1          &  69.8    \\ 
			C     & \ding{51}  & \ding{55}     & \ding{51}   & 69.7          & 70.6      \\ 
			D     & \ding{51}  & \ding{51}  & \ding{51}   & \textbf{70.7}       & \textbf{71.4} \\ \bottomrule[1pt]
		\end{tabular}
		
		\label{tab:table8}
	\end{center}
\end{table}

\subsubsection{Effectiveness of TDMM}
To  evaluate the effectiveness of the proposed TDMM, we first examine its overall contribution. As shown in Table \ref{tab:table5}, integrating the TDMM into the baseline (Task A→D) improves Micro-AUC by $9.9\%$ on UIT-ADrone and Macro-AUC by $10.2\%$ on MUVAD, demonstrating its critical role in capturing fine-grained temporal dynamics and local spatial structures.
We then investigate the sources of these improvements by analyzing its key components. Introducing the STMamba (STM) submodule into the FDSCM-augmented baseline (Task B→C) produces additional gains of $5.8\%$ and $7.2\%$, underscoring the superiority of the Mamba architecture in efficiently modeling complex spatio-temporal dependencies. Furthermore, removing the multi-scale temporal modeling strategy (MST) from the complete model (Task G→C) results in performance drops of $5.4\%$ and $3.8\%$, confirming the necessity of diverse temporal receptive fields.

\begin{figure}[t]
	\centering
	\includegraphics[width=0.45\textwidth,page=1]{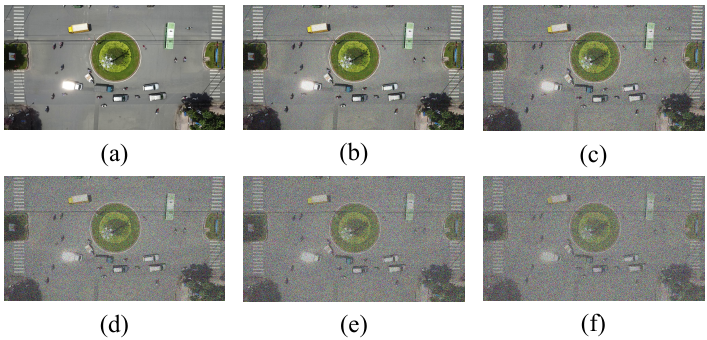} 
	\caption{Visualization of an input frame under varying levels of additive Gaussian noise. (a) is the original frame. The standard deviation of the noise in (b) through (f) is progressively set to 50, 100, 150, 200, and 250.
	}      
	\label{Figure10}
\end{figure}

\begin{table}[t]
	\renewcommand{\arraystretch}{0.95}
    \caption{
    			Performance of FTDMamba on two datasets under additive Gaussian noise with standard deviations from 0 to 250.
		}
        \setlength{\tabcolsep}{7.0pt}
	\begin{center}

		\small
		\begin{tabular}{c| c c c c c c} 
			\toprule[1pt]
			Datasets & 0      & 50 & 100 & 150 & 200 & 250 \\ \midrule
			MUVAD     & \textbf{71.4} & 71.0  & 70.1  & 68.5 & 66.2 & 63.9 \\ 
			UIT-ADrone    & \textbf{70.7} & 69.8 & 69.2 & 67.7 & 65.3 & 63.6  \\ 
			 \bottomrule[1pt]
		\end{tabular}
		
		\label{tab:table9}
	\end{center}
\end{table}

\begin{table}[t]
	\renewcommand{\arraystretch}{0.95}
    \caption{
			PERFORMANCE ON TWO BENCHMARKS WHEN SUBJECTED TO RANDOM FRAME OCCLUSION AT RATIOS FROM 0 TO 0.5.
		}
        \setlength{\tabcolsep}{7.0pt}
	\begin{center}

		\small
		\begin{tabular}{c| c c c c c c} 
			\toprule[1pt]
			Datasets & 0  & 0.1 & 0.2 & 0.3 & 0.4 & 0.5 \\ \midrule
			MUVAD     & \textbf{71.4} & 70.2  & 68.7  & 66.5 & 64.1 & 62.5 \\ 
			UIT-ADrone    & \textbf{70.7} & 70.1 & 69.4 & 68.3 & 66.8 & 64.7  \\ 
			 \bottomrule[1pt]
		\end{tabular}
		
		\label{tab:table10}
	\end{center}
\end{table}

\subsubsection{Effectiveness of TDMM's Structure}
To investigate the optimal internal design of the TDMM module, we analyze the effectiveness of its core scanning strategies, including pixel-wise temporal-first (TF) scanning and patch-wise spatial-first (SF) scanning strategy. 
We compare four different scanning configurations, as shown in Table \ref{tab:table6}. The results clearly demonstrate that our proposed hybrid strategy (Task A) achieves the best performance, reaching AUC scores of $70.7\%$ on UIT-ADrone and $71.4\%$ on MUVAD.  In contrast, single-strategy variants like purely pixel-wise (Task C) or purely patch-wise (Task D) perform less effectively.

We further investigate the impact of STMamba's depth on performance. Table \ref{tab:table7} shows that increasing depth from 1 to 3 (Task A→C) yields marginal accuracy gains. On MUVAD dataset, AUC improves only from $71.4\%$ to $71.7\%$, indicating that a single-layer STMamba sufficiently captures critical spatio-temporal patterns.
deeper modules significantly reduce inference efficiency due to increased computational complexity, with processing speed dropping from 29 FPS to 11FPS, which is inadequate for real-time deployment.
Given the importance of real-time processing in UAV VAD, we set STMamba's depth to 1, balancing detection accuracy with computational efficiency.

\subsubsection{Effectiveness of Parallel Architecture of FDSCM and TDMM} 
To justify our parallel design, we evaluate a cascaded architecture where the FDSCM and TDMM modules are connected sequentially. As shown in Table \ref{tab:table5} (Task H), this configuration leads to  performance degradation, with Micro-AUC scores dropping by $2.0\%$ on UIT-ADrone and $4.2\%$ on MUVAD compared to our parallel framework (Task G). This confirms that the two modules capture complementary rather than sequential information. FDSCM's frequency-domain analysis decouples motion patterns and models global dependencies but sacrifices fine-grained spatial information. The cascaded structure forces TDMM to operate on FDSCM's global representation, creating an information bottleneck. In contrast, our parallel design enables both modules to concurrently process original hierarchical features, integrating global frequency-decoupled context from FDSCM with precise local spatiotemporal details from TDMM for superior performance.

\subsubsection{Effectiveness of Loss Function Combinations}
We assess various combinations of loss functions to validate their effectiveness in optimizing the model. The loss functions include intensity loss ($L_{int}$), gradient loss ($L_{grl}$), and similarity loss ($L_{ssim}$). As shown in Table~\ref{tab:table8}, Task A uses only $L_{int}$ loss, achieving baseline AUC performance of $68.5\%$ and $69.3\%$ on UIT-ADrone and MUVAD. Adding $L_{grl}$ (Task B) or $L_{ssim}$ (Task C) both lead to slight performance improvements. The full combination of all three loss functions (Task D) delivers the best performance. These results demonstrate the complementary nature of the three loss functions and their synergistic effect on model optimization.

\subsection{ Expanded Discussion}
\subsubsection{Robustness Analysis against Input Noise}
To assess FTDMamba's robustness against video quality degradation, we conduct a noise perturbation experiment on the MUVAD and UIT-ADrone datasets. As shown in Fig. \ref{Figure10}, we simulate degradation caused by sensor noise, transmission errors, and compression artifacts by injecting additive Gaussian noise with zero mean and progressively increasing standard deviations $\sigma$ ($\sigma \in \{50,100,150,200,250\}$) into the testing video frames. The quantitative results at different noise levels are presented in Table \ref{tab:table9}. On the MUVAD dataset, the Micro-AUC decreased by only $1.8\%$ when $\sigma=100$. Even under extreme noise conditions ($\sigma=250$), the model maintained a competitive Micro-AUC of $63.9\%$. Besides, a similar degradation pattern was observed on the UIT-ADrone dataset, demonstrating FTDMamba’s robustness to noise perturbations.
\subsubsection{Robustness Analysis against Frame Occlusion}
To evaluate the model's ability to infer anomalies with incomplete temporal information, we design a frame occlusion experiment on the testing datasets of MUVAD and UIT-ADrone. Specifically, we simulate the miss of frames in video sequences by randomly selecting a portion of input frames and setting their pixel values to zero. The occlusion ratio was progressively increased from 0 to $50\%$. As shown in Table \ref{tab:table10}, on the UIT-ADrone dataset, the model's performance decreased by only $3.5\%$ when $30\%$ of the frames are occluded. Even half of the frames are missing, FTDMamba still maintains anomaly detection capabilities.

\section{Conclusion and Future Work}
In this paper, we propose a novel Frequency-Assisted Temporal Dilation Mamba network for UAV VAD following the future frame prediction paradigm. To address the complex multi-source motion coupling problem in dynamic UAV videos, we design a Frequency Decoupled Spatiotemporal Correlation module that effectively separates different motion components and models global spatiotemporal correlations.  Additionally, we design a Temporal Dilation Mamba Module to capture temporal continuity and local spatial correlations at different temporal scales. Furthermore, we construct a large-scale UAV motion video dataset, filling a critical gap in existing datasets which focus on static backgrounds. Extensive experiments validate that our method achieves SOTA performance across three benchmark datasets with both dynamic and static backgrounds.

In future work, we will explore generative models such as diffusion and autoregressive models to explicitly model motion distributions and guide future frame prediction within a probabilistic framework. Meanwhile, given that anomaly definitions depend on scene context and high-level semantics, we will incorporate large language models to provide richer semantic priors, thereby further enhancing the robustness and accuracy of the model’s predictions.

\bibliographystyle{IEEEtran}
\bibliography{reference}

\end{document}